%% file: main.tex
\documentclass[10pt,twocolumn,letterpaper]{article}

\usepackage{cvpr}              

\usepackage{graphicx}
\usepackage{amsmath}
\usepackage{amssymb}
\usepackage{booktabs}
\usepackage{multirow}
\usepackage{booktabs}
\usepackage{algorithm, algorithmic}
\usepackage[table,xcdraw]{xcolor}
\usepackage{marvosym}
\usepackage{makecell}
\makeatletter
\def\thanks#1{\protected@xdef\@thanks{\@thanks
        \protect\footnotetext{#1}}}
\makeatother

\usepackage[pagebackref,breaklinks,colorlinks]{hyperref}

\usepackage[capitalize]{cleveref}
\crefname{section}{Sec.}{Secs.}
\Crefname{section}{Section}{Sections}
\Crefname{table}{Table}{Tables}
\crefname{table}{Tab.}{Tabs.}

\def\cvprPaperID{7903} 
\def\confName{CVPR}
\def\confYear{2023}

\input{math_commands}

\begin{document}

\title{Density-Insensitive Unsupervised Domain Adaption on 3D Object Detection}

\author{Qianjiang Hu\qquad \qquad Daizong Liu\qquad \qquad Wei Hu\textsuperscript{\Letter}
\thanks{\textsuperscript{\Letter} Corresponding author: W. Hu. 
This work was supported by National Natural Science Foundation of China (61972009).}
\\
Wangxuan Institute of Computer Technology, Peking University\\
No. 128, Zhongguancun North Street, Beijing, China\\
{\tt\small hqjpku@pku.edu.cn, dzliu@stu.pku.edu.cn, forhuwei@pku.edu.cn}
}

\maketitle

\begin{abstract}
3D object detection from point clouds is crucial in safety-critical autonomous driving.
Although many works have made great efforts and achieved significant progress on this task, most of them suffer from expensive annotation cost and poor transferability to unknown data due to the domain gap.
Recently, few works attempt to tackle the domain gap in objects, but still fail to adapt to the gap of varying beam-densities between two domains, which is critical to mitigate the characteristic differences of the LiDAR collectors.
To this end, we make the attempt to propose a density-insensitive domain adaption framework to address the density-induced domain gap.
In particular, we first introduce Random Beam Re-Sampling (RBRS) to enhance the robustness of 3D detectors trained on the source domain to the varying beam-density.
Then, we take this pre-trained detector as the backbone model, and feed the unlabeled target domain data into our newly designed task-specific teacher-student framework for predicting its high-quality pseudo labels.
To further adapt the property of density-insensitivity  into the target domain, we feed the teacher and student branches with the same sample of different densities, and propose an Object Graph Alignment (OGA) module to construct two object-graphs between the two branches for enforcing the consistency in both the attribute and relation of cross-density objects.
Experimental results on three widely adopted 3D object detection datasets demonstrate that our proposed domain adaption method outperforms the state-of-the-art methods, especially over varying-density data.
Code is available at \href{https://github.com/WoodwindHu/DTS}{https://github.com/WoodwindHu/DTS}.
\end{abstract}

\vspace{-20pt}
\section{Introduction}
\label{sec:intro}
\input{1_intro}

\vspace{-9pt}
\section{Related Work}
\vspace{-6pt}
\label{sec:related}
\input{2_related}

\vspace{-9pt}
\section{Method}
\vspace{-6pt}
\label{sec:method}
\input{4_method}

\vspace{-9pt}
\section{Experiments}
\vspace{-6pt}
\label{sec:experi}
\input{5_experiments}

\vspace{-8pt}
\section{Conclusion}
\vspace{-6pt}
\label{sec:conclu}
We propose a novel DTS model to bridge the density-induced domain gap for unsupervised domain adaption on 3D object detection. 
In particular, we design Random Beam Re-Sampling to train a density-insensitive detector on the source domain.
To adapt the property of density-insensitivity into the target domain, we then develop a teacher-student framework with Object Graph Alignment to maintain the consistency in both cross-density object attributes and object relations.
Experiments over three datasets demonstrate the effectiveness of the proposed DTS.

{\small
\bibliographystyle{ieee_fullname}
\bibliography{egbib}
}

\end{document}


\title{Supplementary Material for ``Density-Insensitive Unsupervised Domain Adaption on 3D Object Detection''}

\author{Qianjiang Hu\qquad \qquad Daizong Liu\qquad \qquad Wei Hu\textsuperscript{\Letter}
\thanks{
\textsuperscript{\Letter} Corresponding author: W. Hu.}\\
Wangxuan Institute of Computer Technology, Peking University\\
No. 128, Zhongguancun North Street, Beijing, China\\
{\tt\small hqjpku@pku.edu.cn, dzliu@stu.pku.edu.cn, forhuwei@pku.edu.cn}
}

\maketitle


In this supplementary material, we provide more ablation studies and visualizations omitted in our main paper due to the page limit, including 
\begin{itemize}
    \item Section~\ref{sec:additional_ablation}: Additional ablation studies.
    \item Section~\ref{sec:qualitative_results}: Qualitative results.
    \item Section~\ref{sec:performance_under_weekly_supervised_setting}: Performance under weakly supervised setting.
\end{itemize}
As in the main paper, all ablation studies and visualization results in this supplementary file are conducted on the domain adaption case of nuScenes $\rightarrow$ KITTI, using SECOND-IoU as the 3D detection backbone.
\section{Additional Ablation Studies}
\label{sec:additional_ablation}

\paragraph{Sensitivity Analysis of pseudo labels' confidence threshold.} 
As shown in Table~\ref{tab:confidence}, we investigate the effect of different confidence threshold $c_{th}$ in Eq.~(3) of our main paper for pseudo label generation.
We can find that our method achieves the best performance when $c_{th}$ is around $0.6$. 
If $c_{th}$ is even larger, the performance decreases significantly. 
This is because a larger $c_{th}$ gives rise to a smaller number of positive examples that degenerate the self-training process.

\begin{table}[ht]
\centering
\begin{tabular}{c|cc}
\hline
$c_{th}$ & $\text{AP}_{\text{BEV}}$ & $\text{AP}_{\text{AP}{3D}}$ \\ \hline
0.1  & 80.6   & 61.9  \\
0.2  & 80.5   & 64.5  \\
0.3  & 80.2   & 62.8  \\
0.4  & 80.6   & 63.6  \\
0.5  & \textbf{81.4}   & 66.6  \\
0.6  & 81.0   & \textbf{67.2}  \\
0.7  & 71.3   & 59.8  \\
0.8  & 15.6   & 11.7  \\ \hline
\end{tabular}
\caption{Performance under different confidence thresholds $c_{th}$}
\label{tab:confidence}
\end{table}

\paragraph{Sensitivity Analysis of the two terms in Edge-Level Consistency (ELC).} 
Further, we investigate the importance of the two terms in ELC (Eq.~($9$) in the main body): the edge weight alignment and the GLR alignment.
As shown in Table~\ref{tab:gamma}, the performance drops by $2.8$ without the edge weight alignment (\textit{i.e.}, $\gamma=0.0$), and drops by $4.8$ without the GLR alignment (\textit{i.e.}, $\gamma=1.0$). 
This indicates the importance of striking a good balance between the edge weight alignment and the GLR alignment.

\begin{table}[ht]
\centering
\begin{tabular}{c|cc}
\hline
$\gamma$ & $\text{AP}_{\text{BEV}}$ & $\text{AP}_{\text{AP}{3D}}$ \\ \hline
0.0   & 81.2   & 64.9  \\
0.1   & 81.7   & 65.1  \\
0.2   & 81.6   & 63.7  \\
0.3   & 81.6   & 64.6  \\
0.4   & 81.7   & 65.4  \\
0.5   & \textbf{81.4}   & 66.6  \\
0.6   & \textbf{81.4}   & \textbf{67.6}  \\
0.7   & 81.3   & 65.7  \\
0.8   & 81.2   & 63.9  \\
0.9   & 81.0   & 64.1  \\
1.0   & 80.4   & 62.9  \\ \hline
\end{tabular}
\caption{Performance under different $\gamma$}
\label{tab:gamma}
\end{table}

\section{Qualitative Results}
\label{sec:qualitative_results}
\begin{figure*}[h]
    \centering
    \includegraphics[width=\textwidth]{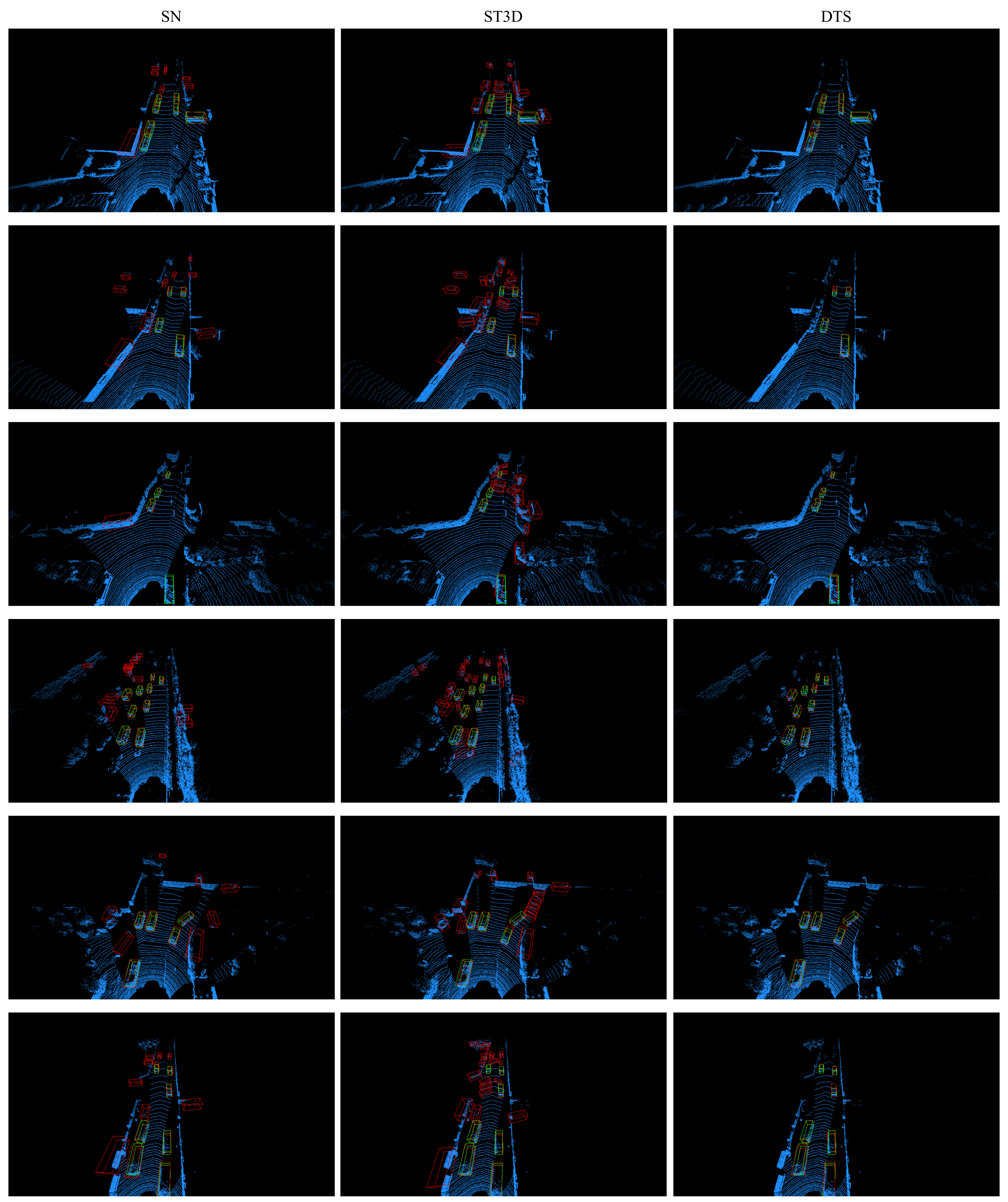}
    \caption{Qualitative results of our proposed DTS and competitive baselines on the KITTI validation set. The green boxes indicate the ground truth bounding boxes, while the red boxes indicate the predicted bounding boxes.}
    \vspace{-10pt}
    \label{fig:visualize_results}
\end{figure*}

\paragraph{Main results.}
As shown in Figure~\ref{fig:visualize_results}, we provide some qualitative results of our proposed DTS and competitive baselines (SN \cite{wang2020train} and ST3D \cite{yang2021st3d}) on the KITTI validation set.
We observe that SN and ST3D produce a few negative predictions, while our predictions are clean and more accurate.
This is because the teacher-student framework with both Node-Level Consistency (NLC) and ELC provides a stable and adaptive pseudo supervision to the detector. 

\begin{figure*}[h]
    \centering
    \includegraphics[width=\textwidth]{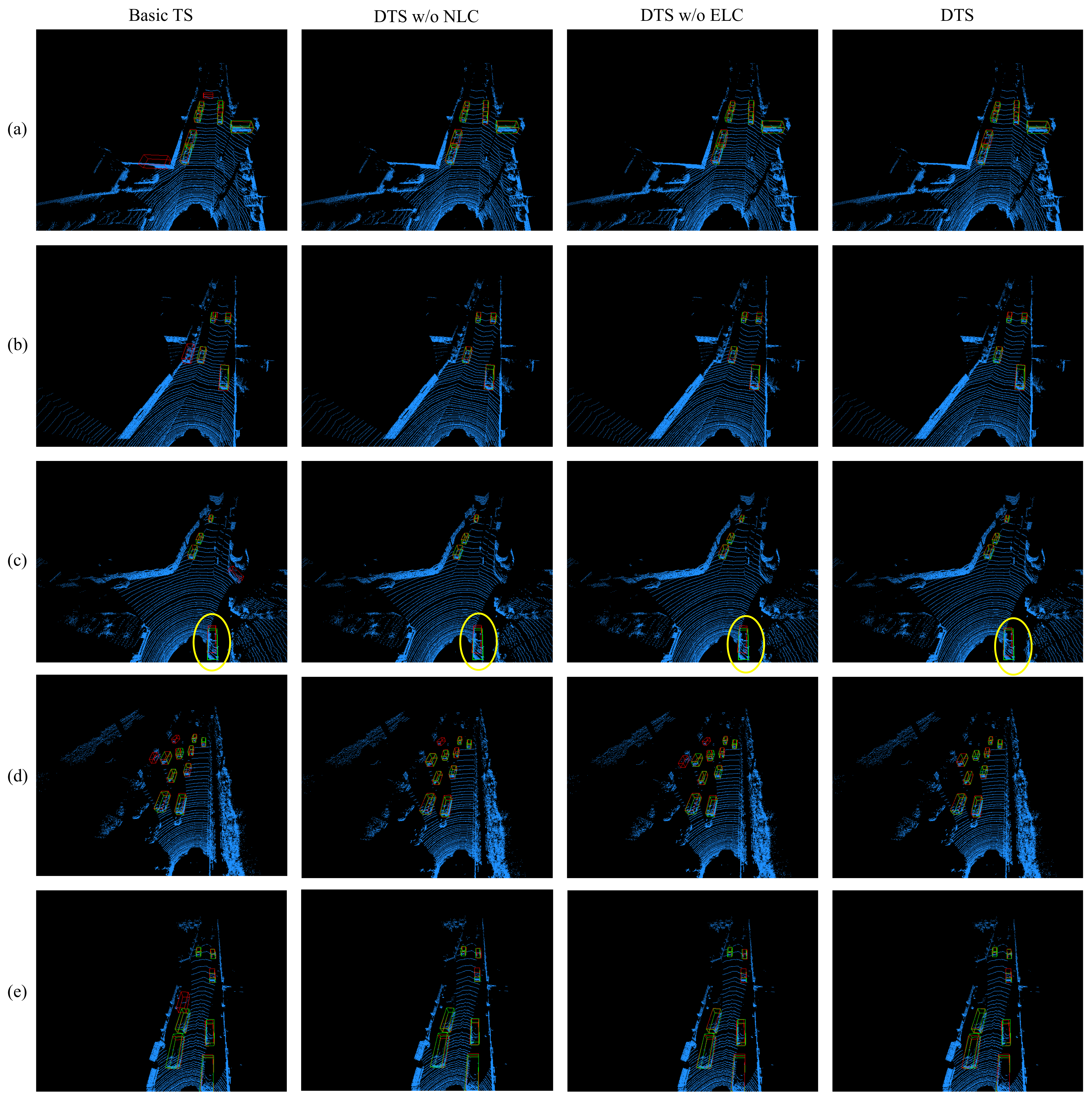}
    \caption{Qualitative results of our proposed DTS and ablation variants. The green boxes indicate the ground truth bounding boxes, while the red boxes indicate the predicted bounding boxes.}
    \vspace{-10pt}
    \label{fig:ablation_results}
\end{figure*}

\paragraph{Ablation results.}
As shown in Figure~\ref{fig:ablation_results}, we also provide some qualitative results of four ablation variants of the proposed DTS: Basic TS (basic teacher-student architecture, {\it i.e.}, DTS without NLC and ELC), DTS without NLC, DTS without ELC, and the complete DTS.
We observe that with NLC and ELC introduced, our DTS reduces the number of negative predictions. 
Also, the complete DTS produces more precise predictions, as clearly demonstrated in regions marked with yellow circles in Figure~\ref{fig:ablation_results}(c).

\section{Performance under weakly supervised setting}
\label{sec:performance_under_weekly_supervised_setting}
Although our method is proposed for UDA, applying additional information (with SN or a few target-domain labels) can further improve the performance, as shown in Table~\ref{tab:supervised}. We observed one needs to  provide around 50\% labels to reach parity with the oracle detector, thus validating the potential applicability.  
\vspace{-6pt}
\begin{table}[h]
    \centering
\scalebox{0.80}{
    \begin{tabular}{l|cc|l|cc}
        \hline
Method                 &  $\text{AP}_{\text{BEV}}$ & $\text{AP}_{\text{3D}}$ & Method                 &  $\text{AP}_{\text{BEV}}$ & $\text{AP}_{\text{3D}}$ \\
\hline 
Ours	               & 81.4   & 66.6 & w/ 20\% label     & 82.4   & 69.5 \\
w/ SN             & 81.4   & 67.0 & w/ 50\% label     & 84.5   & 72.4  \\	
w/ 10\% label     & 81.8   & 67.6 & Oracle            & 83.3   & 73.5\\	
\hline 
\end{tabular}
}
\vspace{-6pt}
\caption{Adaptation performance comparison of unsupervised DA and semi-supervised DA, N $\rightarrow$ K.}
\vspace{-12pt}
\label{tab:supervised}
\end{table}
{\small
\bibliographystyle{ieee_fullname}
\bibliography{egbib}
}

%% file: math_commands.tex
\usepackage{amsmath,amsfonts,bm}

\def\eqref#1{equation~\ref{#1}}

\def\1{\bm{1}}

\def\mF{{\bm{F}}}

\def\mX{{\bm{X}}}
\def\mY{{\bm{Y}}}

\DeclareMathAlphabet{\mathsfit}{\encodingdefault}{\sfdefault}{m}{sl}
\SetMathAlphabet{\mathsfit}{bold}{\encodingdefault}{\sfdefault}{bx}{n}

\def\gE{{\mathcal{E}}}

\def\gG{{\mathcal{G}}}

\def\gL{{\mathcal{L}}}

\def\gN{{\mathcal{N}}}

\def\gS{{\mathcal{S}}}
\def\gT{{\mathcal{T}}}



%% file: 1_intro.tex












3D object detection is a fundamental task in various real-world scenarios, such as autonomous driving \cite{lang2019pointpillars,shi2020pv} and robot navigation \cite{malavazi2018lidar}, aiming to detect and localize traffic-related objects such as cars, pedestrians, and cyclists in 3D point clouds \cite{liu2022imperceptible,hu2022exploring,liu2022point}.
With the advent of deep learning, this task has obtained remarkable advances \cite{lang2019pointpillars,shi2020pv,shi2019pointrcnn,shi2020points,yan2018second,zhou2018voxelnet,yin2021center} in recent years, which however requires costly dense annotations of point clouds.
Further, in real-world scenarios, upgrading LiDARs to other product models can be time-consuming and labor-intensive to collect and annotate massive data for each kind of product, while it is reasonable to use labeled data from previous sensors. 
Also, the number of LiDAR points used in mass-produced robots and vehicles is usually fewer than that in large-scale public datasets \cite{wei2022lidar}. 
To bridge the domain gap caused by different LiDAR beams, it is essential to develop methods that address these differences.
However, the generalization ability of existing methods is proved to be limited \cite{wang2020train} when the 3D models trained on a specific dataset are directly applied to an unknown dataset collected with a different LiDAR, which prevents the wide applicability of 3D object detection in autonomous driving.

\begin{figure}[t]
    \centering
    \includegraphics[width=\columnwidth]{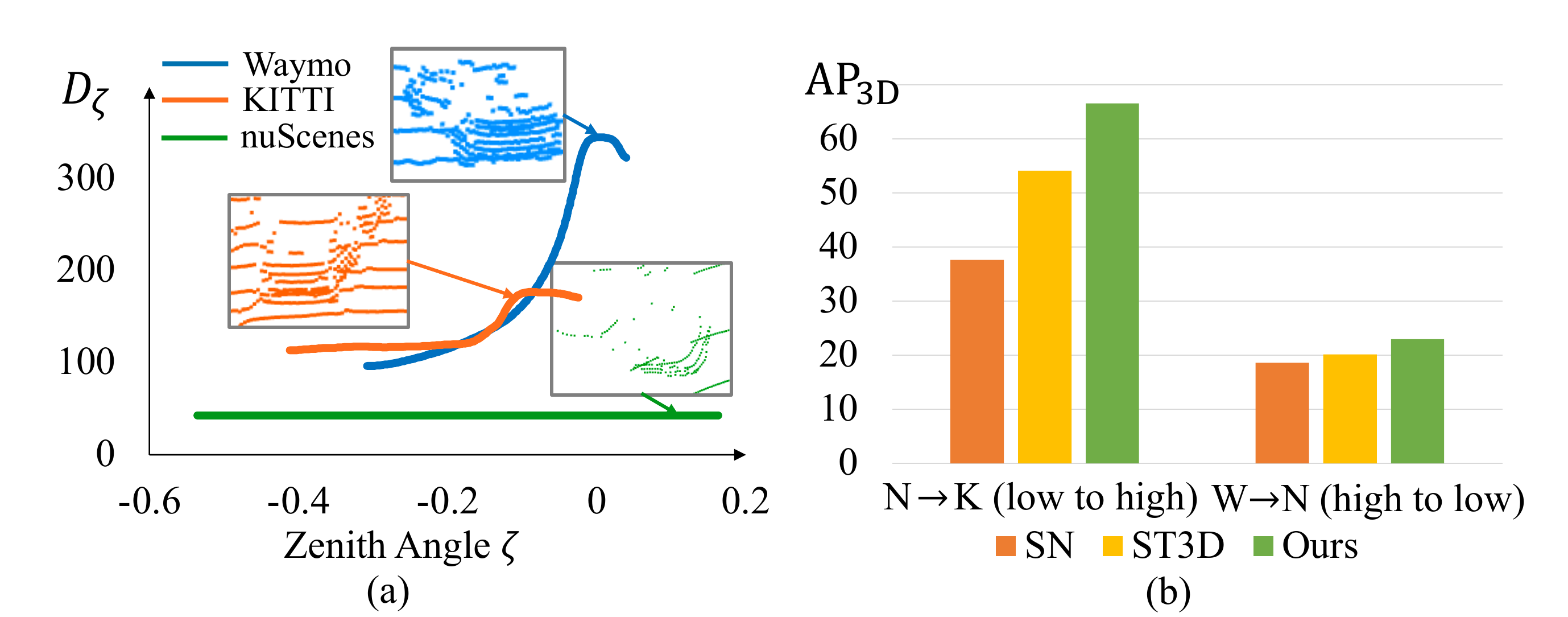}
    \vspace{-16pt}
    \caption{(a) The significant difference of beam densities among Waymo, KITTI, and nuScenes datasets. The beam density $D_\zeta$ represents the number of beams per unit zenith angle. The beams are evenly distributed in nuScenes, while the density of beams in Waymo and KITTI increases as the zenith angle $\zeta$ increases, with the highest density near the horizontal direction. (b) Compared to previous works (SN \cite{wang2020train} and ST3D \cite{yang2021st3d}), our method is more effective in transferring the knowledge from low density to high density or high density to low density (N: nuScenes, K: KITTI, W: Waymo.). }
    \vspace{-20pt}
    \label{fig:teaser}
\end{figure}

To reduce the domain gap between different datasets,
some works \cite{yang2021st3d,luo2021unsupervised,yihan2021learning,wei2022lidar,hegde2021attentive,hegde2021uncertainty,zhang2021srdan,xu2020exploring,wang2020train,saltori2020sf} proposed unsupervised domain adaptation (UDA) methods to transfer knowledge from a labeled source domain to an unlabeled target domain. 
However, most of them focus on reducing the domain gap introduced by the bias in object sizes on the labeled source domain, which neglect another important domain gap induced by {\it varying densities} of point clouds acquired from different types of LiDAR.
We argue that this domain gap is crucial for 3D object detection in two aspects: 
1) As demonstrated in Figure~\ref{fig:teaser}(a), different LiDAR collectors generally produce point cloud data with distinctive densities and distributions, leading to huge density-induced domain gap. 
2) Most 3D detectors are directly trained on a single environment and thus sensitive to the cross-domain density variation. 
As shown in Figure~\ref{fig:teaser}(b), existing domain adaption methods suffer from performance bottlenecks in cross-density scenarios.
Although few works \cite{wei2022lidar} attempt to downsample point clouds of high density and transfer its knowledge to the low-density domain, they are limited to the model design that cannot realize the knowledge transfer from a low-density domain to a high-density domain.
Hence, it is demanded to train robust 3D feature representations that can adapt to point cloud data of varying densities.  

To this end, we make the attempt to propose a novel Density-insensitive Teacher-Student (DTS) framework to address the domain gap induced by varying point densities and distributions.
The key idea of DTS is to first pre-train a density-insensitive object detector on the source domain, and then employ a self-training strategy \cite{zhao2020collaborative,kim2019self,yang2021st3d} to fine-tune this detector on the unlabeled target domain by iteratively predicting and updating its pseudo results.
However, there still remain two concerns:
1) Previous self-training methods may be prone to its mistake by using single-branch prediction. 
2) How to adapt and improve the property of density-insensitivity of the pre-trained 3D detector on the target domain is important.
Therefore, we introduce a task-specific teacher-student framework in order to provide more reliable and robust supervision, in which the teacher and student branches are fed with variants of the same sample in different densities.
Further, considering the object prediction should be invariant in the two branches, we propose to capture their cross-density object-aware consistency for enhancing the density-insensitivity on the target domain.

To be specific, we first introduce Random Beam Re-Sampling (RBRS) to train the density-invariant 3D object detector on the labeled source domain, by randomly masking or interpolating the beams of the point clouds.
Then, we take this pre-trained 3D detector as the backbone model to build a teacher-student framework to iteratively predict and update the pseudo labels on the unlabeled target domain. 
To achieve the goal of density-insensitivity, we feed the student and teacher models with the RBRS-augmented sample and the original sample, respectively.
Moreover, in order to enforce the consistency in attributes and relations of detected objects in the teacher and student branches for more reliable supervision, we construct two graphs based on the objects predicted from the teacher and student models, and propose a novel Object Graph Alignment (OGA) to keep consistent cross-density object-attributes (node-level) and object-relations (edge-level) between the two graphs.
During the training, the student model is optimized based on the predictions of the teacher while the weights of the teacher model are updated by taking the exponential moving average of the weights of the student model. 
In this way, our DTS is effective in reducing the density-induced domain gap and achieving state-of-the-art performance on the unknown target data.

In summary, our main contributions include
\begin{itemize}
\vspace{-8pt}
\item We propose a density-insensitive unsupervised domain adaption framework to alleviate the influence of the domain gap caused by varying density distributions. We develop beam re-sampling to randomize the density of point clouds, which effectively enhances the robustness of 3D object detection to varying densities.
\vspace{-17pt}
\item We exploit a task-specific teacher-student framework to fine-tune the pre-trained 3D detector on the target domain. To adapt and improve the density-insensitivity on the target domain, we introduce an object graph alignment module to keep the cross-density object-aware consistency.
\vspace{-7pt}
\item Experimental results demonstrate our model significantly outperforms the state-of-the-art methods on three widely adopted 3D object detection datasets including NuScenes \cite{caesar2020nuScenes}, KITTI \cite{geiger2013vision}, and Waymo \cite{sun2020scalability}.
\end{itemize}

%% file: 2_related.tex
\paragraph{Point-cloud-based 3D Object Detection:} 
Point-cloud-based 3D object detection \cite{lang2019pointpillars,shi2020pv,yan2018second,shi2019pointrcnn,shi2020points,qi2018frustum,yang2019std,zhou2018voxelnet,chen2017multi,ku2018joint,yang2018pixor,zheng2021se,deng2021voxel} aims to localize and classify objects from point clouds. 
Depending on representation learning strategies, existing works can be divided into three categories: voxel based, point based, and voxel-point based.
Voxel based methods\cite{zhou2018voxelnet,yan2018second,lang2019pointpillars,zheng2021se,deng2021voxel} voxelize point clouds into 2D/3D compact grids and then collapse it to a bird's-eye-view representation.
They are computationally effective but the desertion of fine-grained patterns degrades further refinement.
Point based methods\cite{shi2019pointrcnn,yang20203dssd} directly process point clouds without voxelization.
These methods wholly preserve the irregularity and locality of a point cloud but have relatively higher latency.
Point-voxel based methods\cite{yang2019std,shi2020pv} integrate the advantages of both voxel based methods and point based methods together. 
Following previous works \cite{yang2021st3d,wei2022lidar}, we adopt voxel based PointPillars \cite{lang2019pointpillars}, SECOND \cite{yan2018second} and point-voxel based PV-RCNN\cite{shi2020pv} as our detectors.

\begin{figure*}[h]
    \centering
    \includegraphics[width=0.83\textwidth]{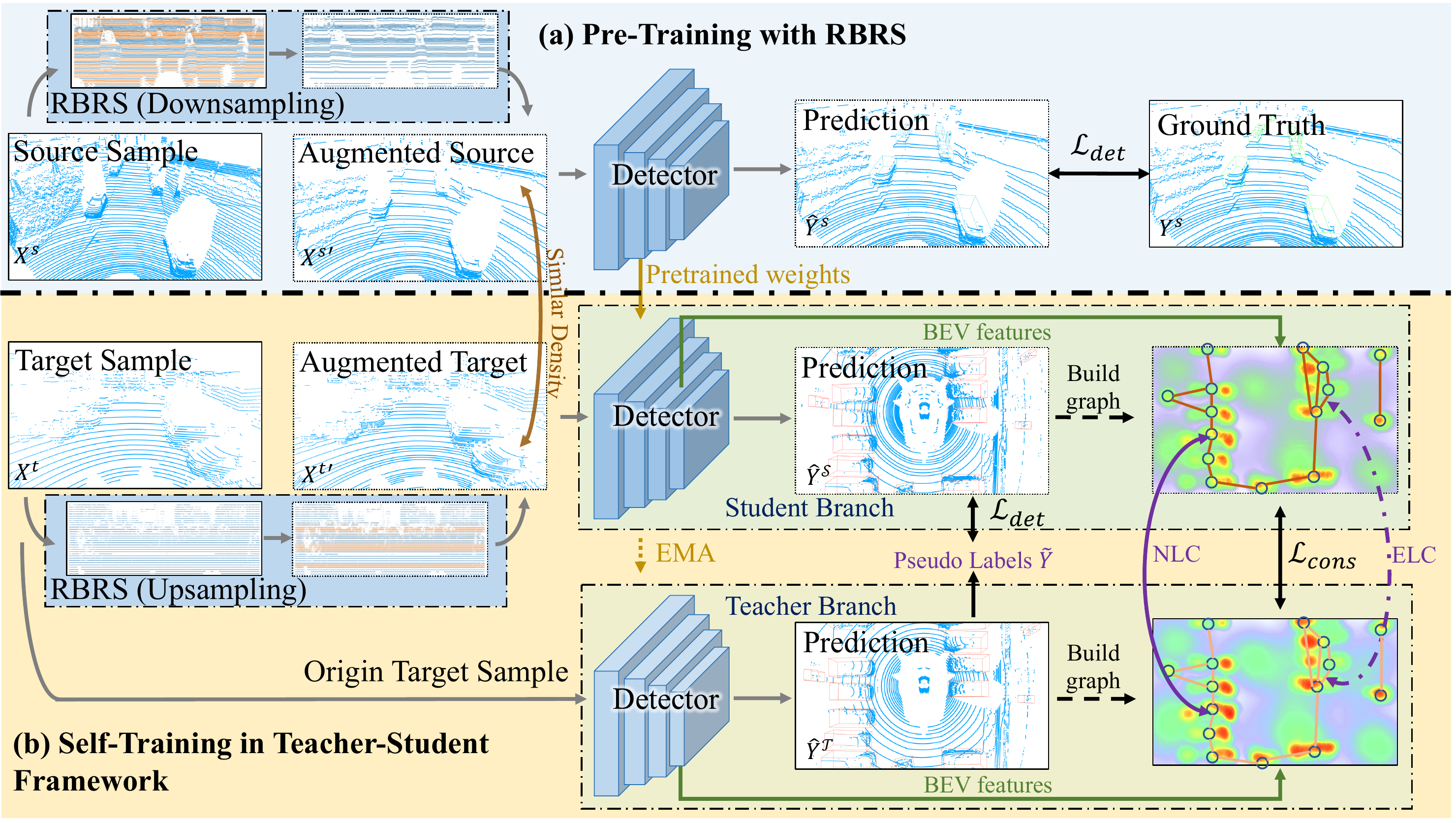}
    \caption{{ The overall pipeline of the proposed DTS. Here, we take the case of transferring the high-density domain into low-density domain as an example.} Given the labeled source data and the unlabeled target data with different point density distributions, (a) our DTS first pre-trains a 3D detector on the source data with random beam re-sampling (RBRS), where the Yellow beams mark the masked/interpolated beams; (b) then DTS builds a teacher-student self-training framework with the pre-trained 3D detector. During the self-training process, the student and teacher branches are fed with different density variants of the same input. An object graph alignment module is further deployed to capture the cross-density object-aware consistency between the two branches.}
    \vspace{-10pt}
    \label{fig:framework}
\end{figure*}
\vspace{-12pt}
\paragraph{Unsupervised Domain Adaptation for 2D/3D Object Detection:} 
UDA aims to transfer the model trained on the fully annotated source domain to the unannotated target domain. 
A variety of solutions \cite{zhao2020collaborative,kim2019self,khodabandeh2019robust,roychowdhury2019automatic,saito2019strong,chen2018domain,hsu2020every,sindagi2020prior,vs2021mega,rodriguez2019domain,kim2019diversify,cai2019exploring,deng2021unbiased} have been proposed in the 2D object detection task.
As the pioneer, Ben \textit{et al.}\cite{ben2010theory} designed a $\mathcal{H}\Delta\mathcal{H}$-distance to measure the divergence between two domains that have different data distributions and proposed a general framework to perform domain adaptation. 
Inspired by GAN\cite{goodfellow2020generative}, many methods in the literature\cite{saito2019strong,chen2018domain,hsu2020every,sindagi2020prior,vs2021mega} generate a domain discriminator to correctly classify the source/target domain while training a detector model to fool the domain discriminator.
Some other methods\cite{rodriguez2019domain,kim2019diversify} follow the domain randomization strategy to devoid all source-style bias on the source detector. 

While a lot of research has been conducted on UDA for object detection with 2D image data, there is relatively little literature in the field of UDA for 3D object detection. 
Wang \textit{et al.} proposed statistical normalization (SN) \cite{wang2020train} to normalize the object sizes of the source and target domains so they could bridge the domain gap introduced by the difference in object sizes. 
SPG \cite{xu2021spg} utilized the semantic point generation to tackle the domain gaps induced by deteriorated point cloud quality.
SF-UDA \cite{saltori2020sf} uses the temporal coherency to estimate the object size in the target domain while getting rid of the target domain statistics.
3D-CoCo \cite{yihan2021learning} explores a contrastive co-training framework including separate 3D encoders to provide more stable supervisions from the labeled source data while avoiding the biased knowledge of the source domain.
MLC-Net \cite{luo2021unsupervised} implements a mean-teacher paradigm and exploits the point-, instance- and neural statistics-level consistency to facilitate the cross-domain transfer.
Yang \textit{et al.} proposed ST3D \cite{yang2021st3d} which redesigns the self-training pipeline to improve the quality of pseudo-labels for 3D object detection. 
Although these methods successfully achieve performance improvement compared to direct transfer, they neglect the density-induced domain gap.
Observing that ST3D is hard to adapt detectors from data with more beams to data with fewer beams, Wei \textit{et al.} proposed LiDAR Distillation \cite{wei2022lidar}, which downsamples the high-density data to align the point cloud density of the source and target domains. 
Then they progressively distill the knowledge from the high-density data to the low-density data. 
Different from LiDAR Distillation that is limited to transferring from high density to low density, our proposed DTS is able to transfer knowledge under various settings of point densities. 

%% file: 4_method.tex
\subsection{Problem Statement and Overview}
\vspace{-6pt}
\label{subsec:overview}
Unsupervised domain adaption for 3D object detection aims to transfer a model trained on a labeled source domain to an unlabeled target domain. 
Generally, the source domain is a point cloud dataset $\{\mX_i^s\}_{i=1}^{N_s}$ labeled with the corresponding class $a_s$ and bounding box $\{\mY_i^s\}_{i=1}^{N_s}$, while the target domain is an unlabeled point cloud dataset $\{\mX_i^t\}_{i=1}^{N_t}$, where $s$ and $t$ represent source and target domains respectively, and $i$ means the $i$-th instance. 
$N_s$ and $N_t$ are the number of source and target point clouds, respectively.
Generally, the label $Y_i^s$ is a seven-dimensional vector, which is parameterized by its center location $\mathbf{c}=\{c_x, c_y, c_z\}$, bounding box size $\mathbf{b}=\{l,w,h\}$, and the yaw angle $\xi$, respectively.
Since point clouds of different domains are often collected by different LiDAR equipments with various density distributions, it is demanded to develop a density-insensitive domain adaptation model to reduce this density-induced domain gap.



To this end, we propose a novel Density-insensitive Teacher-Student (DTS) framework, as shown in Figure \ref{fig:framework}, which mainly consists of three components.
\begin{itemize}
\vspace{-7pt}
\item Given the source and target domain data, DTS first pre-trains a density-insensitive 3D detector on the source data with Random Beam Re-Sampling (RBRS) to reduce the density-induced domain gap.
\vspace{-7pt}
\item Then the pre-trained 3D detector is taken as the backbone to build a task-specific teacher-student self-training framework, in which the teacher model generates the pseudo labels of the target data and provides high-quality supervision to update network weights of the student model. 
To adapt the property of density-insensitivity on the target domain, we feed the student and teacher models with the RBRS-augmented and the original target samples, respectively.
\vspace{-7pt}
\item We further propose an Object Graph Alignment (OGA) module to capture and learn the cross-density object-aware consistency between student and teacher models for improving the density-insensitivity.
\end{itemize}
\vspace{-6pt}
In the following, we provide the details of each component.

\vspace{-6pt}
\subsection{Pre-training with Random Beam Re-Sampling}
\vspace{-6pt}
\label{subsec:pretrain}
Before transferring the knowledge from the source domain into the target domain, we need to pre-train a 3D detector to extract a wealth of knowledge on the annotated source data $\{(\mX_i^s,\mY_i^s)\}_{i=1}^{N_s}$. 
However, this learned knowledge contains the domain-specific bias due to different object sizes and point densities collected by different LiDAR, leading to the poor generalization ability on the target domain.
Although Yang \textit{et al.}\cite{yang2021st3d} seek to overcome the bias in object sizes via random object scaling (ROS), the density bias has been seldom investigated.
Therefore, we propose to reduce this density gap among different domains and provide a density-insensitive 3D detector.

Given point clouds collected with $M$-beam LiDAR, we first denote the zenith angle of the $j$-th beamas as $\zeta(j)$.
The density of beams, \textit{i.e.}, the count of beams in the unit zenith angle, could be approximated by $D_{\zeta}(j) = 1/(\zeta(j+1)-\zeta(j))$.
Inspired by the success of ROS in overcoming the bias in object sizes, we propose random beam re-sampling (RBRS), a simple yet effective strategy, to train a \textit{density-insensitive} 3D detector.

\vspace{-14pt}
\paragraph{Random Beam Re-Sampling.} 
RBRS aims to randomly down-sample the dense data and up-sample the sparse data.
Before re-sampling, we first transfer cartesian coordinates $(x,y,z)$ of points to the spherical coordinates as:
\vspace{-6pt}
\begin{equation}
\vspace{-4pt}
\begin{aligned}
\zeta &=\arctan \frac{z}{\sqrt{x^2+y^2}}, \\
\phi &=\arcsin \frac{y}{\sqrt{x^2+y^2}}, \\
r &= \sqrt{x^2+y^2+z^2},
\end{aligned}
\end{equation}
where $\zeta$ and $\phi$ are zenith and azimuth angles, $r$ is the distance from each point to the LiDAR sensor.
By taking the zenith angle as the vertical coordinate and the azimuth angle as the horizontal coordinate, point clouds could be transformed to range images (like RBRS blocks in Figure~\ref{fig:framework}). 
We use the K-Means algorithm \cite{jain1988algorithms,macqueen1967classification} to mark the the ID of beams in the range images according to its zenith angle $\zeta$.
Then we perform RBRS by randomly down-sampling or up-sampling the range images and reverse them into point clouds.

Specifically, to down-sample the dense data, for the $j$-th beam with the beam density of $D_{\zeta}(j)$, RBRS randomly masks this beam with the probability of $\eta_j$ according to the beam density $D_{\zeta}(j)$.
Considering a beam with a larger beam density is more likely to be masked, we formulate this mask probability $\eta_j$ as $\eta_j = 1- \gamma_1 / {D_{\zeta}(j)}$,
where $\gamma_1$ is a factor to control the overall density of the sampled point cloud.

To up-sample the sparse data, RBRS randomly interpolates artificial beams between the original beams.
Specifically, an artificial beam is interpolated with a probability of $\eta_j'$ between the $j$-th beam and the $(j+1)$-th beam.
Considering a beam with a larger $D_{\zeta}(j)$ has smaller $\eta_j'$,
, the interpolation probability $\eta_j'$ is set as $\eta_j' = \gamma_2 / {D_{\zeta}(j)}$,
where $\gamma_2$ is the factor, and a larger $\gamma_2$ indicates more beams to be interpolated.
Then, if a new beam is to be interpolated between the $j$-th beam and the $(j+1)$-th beam, every point in the $j$-th beam is selected as datum mark.
Taking a point $k$ in the $j$-th beam as an example, assuming its spherical coordinate is $(\zeta_k, \phi_k, r_k)$, we first find a point in the $(j+1)$-th beam, marked as $k'$, which is closest to point $k$ measured in the azimuth angle. 
Then the spherical coordinate of the newly interpolated point is defined as:
\vspace{-9pt}
\begin{equation}
\vspace{-6pt}
\begin{aligned}
    \zeta =\frac{\zeta_k+\zeta_{k'}}{2}, \quad
    \phi =\frac{\phi_k+\phi_{k'}}{2}, \quad
    r = \frac{r_k+r_{k'}}{2}.
\end{aligned}
\end{equation}
Subsequently, the newly interpolated point is transformed back to Cartesian coordinates for concatenating with the original points. At last, we utilize this RBRS strategy to pre-train the 3D detector on the source data.


\vspace{-6pt}
\subsection{Our Basic Teacher-Student Architecture}
\vspace{-6pt}
\label{subsec:basic_teacher_student}
After obtaining the pre-trained density-insensitive 3D detector on the source domain, we take it as the backbone model to build a self-training framework for fine-tuning the source domain knowledge into the target domain.
Motivated by the success of the teacher-student paradigm \cite{tarvainen2017mean,cubuk2018autoaugment,berthelot2019mixmatch,xie2020unsupervised} in semi-supervised learning, we design a task-specific teacher-student paradigm to fine-tune the 3D detector on the target domain.

Specifically, the teacher-student framework consists of two separate branches, \textit{i.e.}, a non-trainable teacher detector $D_{\mathcal{T}}$ and a trainable student detector $D_{\mathcal{S}}$, sharing the same architecture.
For each input $\mX_i^t$ of the unlabeled target domain, the teacher detector is fed with its original data $\mX_i^t$, while the student detector is fed with the augmented version $\mX_i^{t'}$ via our RBRS.
This operation is to adapt the property of density-insensitivity of the source domain into the target domain.
We denote the generated bounding boxes of the teacher branch and the student branch as $\hat{\mY_i}^{\mathcal{T}}$ and $\hat{\mY_i}^{\mathcal{S}}$, respectively.

\begin{figure}[t]
    \centering
    \includegraphics[width=0.9\columnwidth]{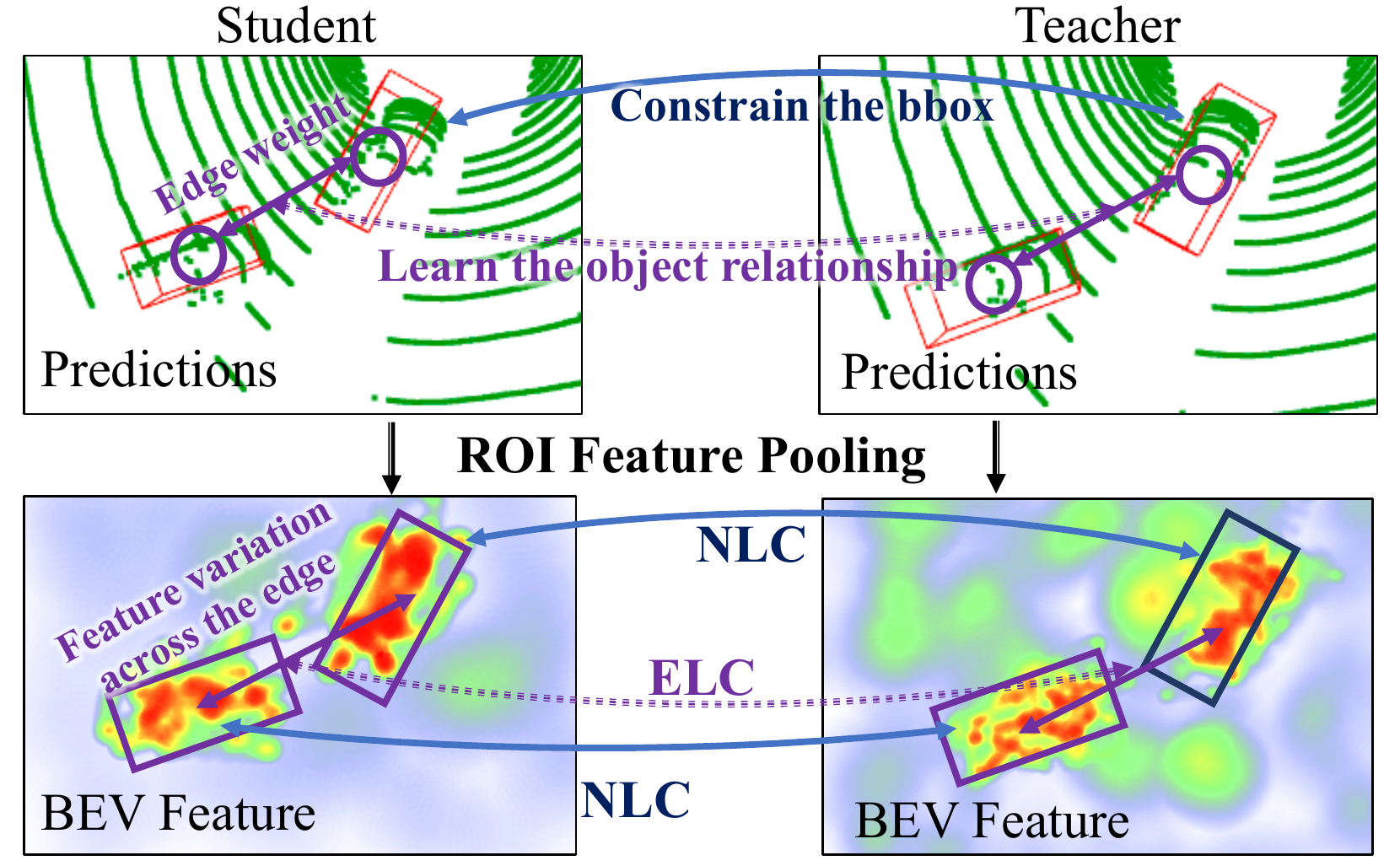}
    \vspace{-10pt}
    \caption{Illustration of our proposed Node-Level Consistency (NLC) and Edge-Level Consistency (ELC). We maintain the NLC to constrain the bounding-box regression and utilize the ELC to learn the object relations for better discriminating the objects.}
    \vspace{-15pt}
    \label{fig:zenith}
\end{figure}

During the teacher-student framework learning, the teacher detector is used to generate pseudo labels of the input $\mX_t$ and provide the supervision signal to train the student detector.
In particular, the predictions of the teacher branch with confidence higher than a threshold $c_{th}$ are chosen to generate the pseudo labels as:
\vspace{-9pt}
\begin{equation}
\vspace{-9pt}
    \tilde{\mY_i} = \{\hat{\mathbf{y}}_j \in \hat{\mY_i}^{\mathcal{T}}|c_j > c_{th}\},
    \label{eq:pseudo_label}
\end{equation}
where $\hat{\mathbf{y}}_j$ is the $i$-th predicted bounding box in $\hat{\mY_i}^{\mathcal{T}}$ and $c_j$ is the confidence of $\hat{\mathbf{y}}_j$.
The student detector takes these pseudo labels for supervision as:
\vspace{-9pt}
\begin{equation}
\vspace{-9pt}
    \mathcal{L} = \mathcal{L}_{\text{det}} + \mathcal{L}_{\text{cons}},
    \label{eq:total_loss}
\end{equation}
where $\mathcal{L}_{\text{det}}$ is the detection loss of the target-domain samples and is supervised by the pseudo label $\tilde{\mY_i}$, and $\mathcal{L}_{\text{cons}}$ is the consistency loss between the teacher and the student branch, which will be elaborated in Section~\ref{subsec:consistency}.

To further improve the quality of the pseudo labels, we also deploy the exponential moving average (EMA) technique \cite{tarvainen2017mean} to update the weight of the teacher detector as:
\vspace{-9pt}
\begin{equation}
\vspace{-4pt}
    \theta^{\mathcal{T}} = \alpha \theta^{\mathcal{T}} + (1-\alpha) \theta^{\mathcal{S}},
    \label{eq:ema}
\end{equation} 
where $\alpha$ is a smoothing coefficient hyperparameter.
The moving average in Eq.~\ref{eq:ema} makes $\theta^{\mathcal{T}}$ evolve more smoothly than $\theta^{\mathcal{S}}$.
As a result, the teacher can aggregate information after every step and generate stable predictions of the input.

\subsection{Object-Graph Consistency Learning between the Teacher and Student Branches}
\label{subsec:consistency}
To further enhance the property of density-insensitivity in the target domain, we propose to capture the cross-density consistency among the student and teacher branches.
Considering the detected objects of different-density variants of the same sample should be invariant, we build a contextual graph based on the predicted objects of each branch for aligning both node-level (object-attribute) and edge-level (object-relation) information between the two branches, as shown in Figure~\ref{fig:zenith}.
Different from MLC-Net which also explores the multi-level consistency, we enforce consistency from objects’ relations for capturing their similarity, by building graphs to model the object-level consistency via NLC and the \textbf{relation-level consistency} via ELC, which considers both local and global features.

\vspace{-16pt}
\paragraph{Object Graph Construction} 
\label{subsubsec:graph}
According to the predicted objects of each branch, we construct a fully-connected undirected graph $\gG=\{\gN, \gE\}$ to model the object relationship.
For the graph of each branch,
$\gN =  \{\mathbf{y}_i \in \hat{\mY_t}^{\mathcal{T}}|c_i > c_{th_{\gG}}\}$ is the node set where each node represents an object prediction.
$\gE$ is the edge set where each edge represents the relationship between the connected objects.
Here, we utilize a heuristic method to weigh the edges between the objects by 1) the close-distance objects have a stronger connection; 2) the objects share similar sizes have a stronger connection; 3) the objects in the same direction indicate they are driving in the same direction, thus they have a stronger connection.
Based on these, we set the edge weight corresponding to the above connected objects' location $\mathbf{c}$, bounding box size $\mathbf{b}$ and yaw angle $\xi$ as:
\vspace{-9pt}
\begin{equation}
\vspace{-4pt}
    w_{ij} = \exp\left(-\frac{||\mathbf{c}_i - \mathbf{c}_j||_2^2 + \epsilon_1||\mathbf{b}_i -  \mathbf{b}_j||_2^2 + \epsilon_2 (\xi_i - \xi_j)^2}{\tau^2}\right),
    \label{eq:edge}
\end{equation}
where $\epsilon_1$ and $\epsilon_2$ control the importance of the object size and yaw angle, and $\tau$ is a temperature hyperparameter. $||\cdot||_2$ is the L2 norm operation.
We denote the graphs of the teacher and the student as $\gG^{\gS}=\{\gN^{\gS}, \gE^{\gS}\}$ and $\gG^{\gT}=\{\gN^{\gT}, \gE^{\gT}\}$, respectively.

\begin{algorithm}[t!]
\renewcommand{\algorithmicrequire}{\textbf{Require:}}
\renewcommand{\algorithmicensure}{\textbf{Output:}}
\caption{Training Process of our DTS}
\label{alg:dts}
\begin{algorithmic}[1]
\REQUIRE 
Labeled source domain data $\{\mX_i^s, \mY_i^s\}_{i=1}^{N_s}$, and unlabeled target domain data $\{\mX_i^t\}_{i=1}^{N_t}$
\STATE Pre-train the object detector on $\{\mX_i^s, \mY_i^s\}_{i=1}^{N_s}$ with RBRS as detailed in Sec.~\ref{subsec:pretrain}.
\STATE Take the pre-trained model as the backbone to build the teacher-student architecture.
\FOR{$i=1$ to $N_t$}
\STATE Forward the teacher-student network demonstrated in Figure~\ref{fig:framework} with $\mX_i^t$.
\STATE Generate pseudo labels $\tilde{\mY}_i$ using the teacher's output $\hat{\mY}_i$ with Eq.~\ref{eq:pseudo_label}.
\STATE Calculate the bounding-box supervision $\gL_{det}$ of the student branch with $\tilde{\mY}_i$.
\STATE Construct object graphs over the two branches as detailed in Sec.~\ref{subsubsec:graph}
\STATE Calculate the cross-graph node-level consistency loss $\gL_{node}$ and edge-level consistency loss $\gL_{edge}$ as detailed in Sec.~\ref{subsubsec:nlc}
\STATE Back-forward the total loss in Eq.~\ref{eq:total_loss} to update the student network.
\STATE Update the teacher's weight using Eq.~\ref{eq:ema}.
\ENDFOR
\STATE Go back to Line 3 until convergence.
\ENSURE The object detection model for the target domain.
\end{algorithmic}  
\end{algorithm}

\vspace{-16pt}
\paragraph{Node-Level Consistency} 
\label{subsubsec:nlc}
Since the inputs of both teacher and student branches are generated from the same sample with different densities, their object predictions should be density-invariant, \textit{i.e.}, $\gN^{\gS}$ and $\gN^{\gT}$ is fully matched.
Therefore, we aim to capture the node-level consistency between the graphs of two branches for aligning the predicted bounding boxes of the same object from inputs of different densities.

Specifically, we first calculate the IoU between the student predictions $\gN^{\gS}$ and the teacher prediction $\gN^{\gT}$.
If the IoU is larger than a threshold $IoU_{th}$, we consider these object detection results are matched and belong to the same object.
The detection results with lower IoU are filtered out. 
For each matched object, we utilize its bird's eye view (BEV) features sampled via region of interest (ROI) as the guidance to pull its predicted bounding boxes of two branches closer. 
Considering that the matched object features of the two branches should be similar, we design the node-level consistency loss $\mathcal{L}_{\text{node}}$ to maximize their similarity in attributes as:
\vspace{-9pt}
\begin{equation}
\vspace{-7pt}
    \mathcal{L}_{\text{node}} = \frac{1}{N} \sum_{i=1}^N \exp\left(-\frac{({\mathbf{f}_i^{\mathcal{S}}})^{\top}  \mathbf{f}_i^{\mathcal{T}}}{||\mathbf{f}_i^{\mathcal{S}}||_2  ||\mathbf{f}_i^{\mathcal{T}}||_2}\right),
\end{equation}
where $N$ is the number of the matched object pairs, $\mathbf{f}_i^{\mathcal{S}}\in R^{C}$ and $\mathbf{f}_i^{\mathcal{T}} \in R^{C}$ are the $C$-channel ROI features of the $i$-th object from the student and the teacher detectors.

\vspace{-16pt}
\paragraph{Edge-Level Consistency}
\label{subsubsec:elc}
In addition to the object-level consistency, the edge-level consistency is also worth exploring since the same object in two branches should contribute similarly to its neighboring objects.
To align the edges of the graphs in two branches, we consider to not only match the corresponding edge weights but also constrain the same feature variation across the edges.
Specifically, the edge weight alignment could be implemented by minimizing the difference between the teacher's edge weight and the student's edge weight.
Besides, to measure the feature variation across the edges, we introduce the graph Laplacian regularization (GLR) \cite{ando2006learning,hu2021graph} as:
\vspace{-8pt}
\begin{equation}
\vspace{-8pt}
    GLR = \mathbf{tr}(\mF^T(\mathbf{D} - \mathbf{W}) \mF),
\end{equation}
where $\mathbf{tr}(\cdot)$ calculates the trace of the matrix, $\mF \in R^{N \times C}$ is the concatenated features of nodes, $\mathbf{W} \in R^{N \times N}$ is the adjacency matrix with each entry denoting an edge weight $w_{i,j}$, $\mathbf{D}$ is the degree matrix---a diagonal matrix where $ d_{ij} = \sum_{j=1}^N w_{ij} $.
GLR measures the smoothness of features with respect to the graph: a smaller GLR represents smaller variation in node features across edges.
Hence, we define the edge-level consistency loss $\mathcal{L}_{\text{edge}}$ as:
\vspace{-9pt}
\begin{equation}
\vspace{-4pt}
    \mathcal{L}_{\text{edge}} = \frac{1}{N^2} \left( \gamma ||\mathbf{W}^{\mathcal{T}} - \mathbf{W}^{\mathcal{S}} ||_2^2 + (1-\gamma) \mathbf{tr}(GLR^\gS - GLR^\gT)\right),
\end{equation}
where $\gamma$ strikes a balance between the two terms. 
The first term represents the difference in edge weights between the two graphs, while the second term aims to enforce the feature variation of the student branch to be similar with that of the teacher branch.

Overall, the joint consistency loss $\mathcal{L}_{\text{cons}}$ is defined as:
\vspace{-9pt}
\begin{equation}
\vspace{-6pt}
    \mathcal{L}_{\text{cons}} = \beta_1 \mathcal{L}_{\text{node}} + \beta_2 \mathcal{L}_{\text{edge}},
\end{equation}
where $\beta_1$ and $\beta_2$ are the hyperparameters to control the involvement of the two consistency losses. The training process of the whole DTS is summarized in Algorithm~\ref{alg:dts}.

%% file: 5_experiments.tex
\subsection{Experimental Settings}
\input{tables/datasets}
\input{tables/main_experiment}
\vspace{-6pt}
\paragraph{Datasets.} 
We conduct experiments on three widely used autonomous driving datasets: KITTI \cite{geiger2013vision}, Waymo \cite{sun2020scalability}, and nuScenses \cite{caesar2020nuScenes}. 
The KITTI\cite{geiger2013vision} contains 7481 frames of point clouds for training and validation, and all the data is collected with 64-beam Velodyne LiDAR.
The Waymo \cite{sun2020scalability} dataset contains 122000 training and 30407 validation frames of point clouds collected with five LiDAR sensors, \textit{i.e.}, one 64-beam LiDAR and four 200-beam LiDAR. 
The nuScenes dataset \cite{caesar2020nuScenes} contains 28130 training and 6019 validation point clouds collected with a 32-beam roof LiDAR.
Table~\ref{tab:datasets} shows an overview of the three datasets. Note that there is a huge difference in their densities (also shown in Figure~\ref{fig:teaser}(a)).
Following previous works \cite{yang2021st3d, yang2021st3d++}, we evaluate our DTS by adapting across domains with different LiDAR-beam densities (Waymo $\rightarrow$ nuScenes, Waymo $\rightarrow$ KITTI and nuScenes $\rightarrow$ KITTI).

\vspace{-11pt}
\paragraph{Evaluation metrics.}
We follow \cite{yang2021st3d} and adopt the KITTI evaluation metric for evaluating our methods on the commonly used car category (the vehicle category in Waymo). 
In detail, we use the average precision (AP) as the evaluation metric for both BEV IoUs and 3D IoUs under an IoU threshold of 0.7 over 40 recall positions.
We also adopt the domain adaptation metric (\textit{i.e.}, Closed Gap) \cite{yang2021st3d} to demonstrate the effectiveness on domain adaption, which is defined as $Closed\ Gap = \frac{\text{AP}_{model}\ -\ \text{AP}_{source}}{\text{AP}_{oracle}\ - \text{AP}_{source}}\times 100\%$.

\vspace{-11pt}
\paragraph{Implementation details.}
We validate the proposed DTS on three detection backbones SECOND-IoU\cite{yang2021st3d}, PV-RCNN\cite{shi2020pv} and PointPillars\cite{lang2019pointpillars}. 
We adopte the training setup of the popular point cloud detection codebase open pcdet\cite{openpcdet2020} to pre-train our detectors on the source domain.
For the following target domain self-training stage, we use Adam\cite{kingma2014adam} and one cycle scheduler to fine-tune the detectors for 30 epochs.
The learning rate is set to $1.5\times10^{-3}$.
The EMA smoothing coefficient hyperparameter $\alpha$ is set to $0.999$.
The confidence threshold $c_{th}$ of pseudo labels as well as the confidence threshold $c_{th_{\gG}}$ for node construction are both set to $0.5$.
We set $\lambda_1$ and $\lambda_2$ to $5.0$ and $20.0$ empirically, which control the participation of object sizes and yaw angles to build object graphs.
The temperature $\tau$ is set to $13.0$.
The involvement hyperparameters of the two consistency losses $\beta_1$ and $\beta_2$ are set to $0.05$ and $0.3$ respectively, while $\gamma$ is set to $0.5$.
The threshold to match nodes $IoU_{th}$ is set to $0.1$.
We set the beam interpolation probability factor as $\epsilon_2=25.0$ to up-sample nuScenes and the beam mask probability factor as $\epsilon_1=75.0$ and $\epsilon_1=100.0$ to down-sample KITTI and Waymo respectively. 

\vspace{-9pt}
\subsection{Comparison with State-of-the-Arts}

\vspace{-3pt}
\paragraph{Main Results.}
We compare our proposed DTS with SN\cite{wang2020train}, ST3D\cite{yang2021st3d}, ST3D++\cite{yang2021st3d++}, 3D-CoCo\cite{yihan2021learning} and L.D.\cite{wei2022lidar}.
As shown in Table~\ref{table:main}, DTS outperforms all compared methods by large margins on all domain adaptation settings.
SN, ST3D and ST3D++ overcome the bias of object sizes in the source domain effectively, however,
their ignorance of the density-induced domain gap sacrifices some of their performance.
3D-CoCo includes separate 3D encoders for the source and target data, making it hard to utilize the useful knowledge from the 3D encoder of the source branch, thus leading to worse performance.
Compared to L.D. that only transfers knowledge from high density into low density, our DTS is more density-insensitive.

Further, while our method is proposed for UDA, we observed one needs to provide around 50\% labels to reach parity with the oracle detector, thus validating the potential applicability (Supplementary Sec.~\textcolor{red}{S3}).

\begin{figure}[t!]
    \centering
    \includegraphics[width=0.8\columnwidth]{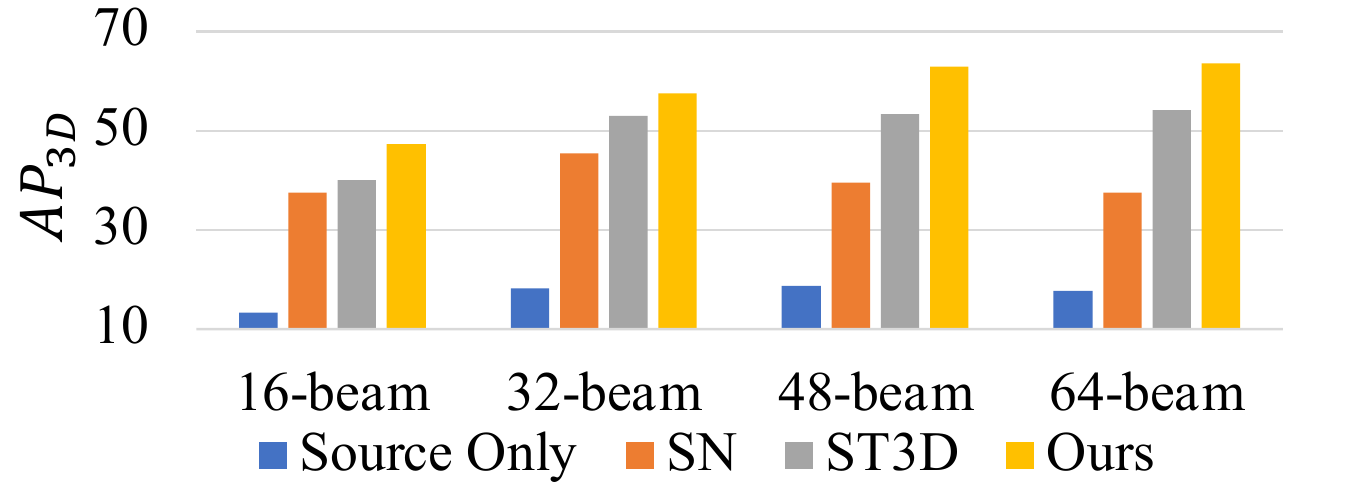}
    \vspace{-10pt}
    \caption{Adapting the model trained on nuScenes data to different down-sampled KITTI data of different densities.}
    \label{fig:different_beam}
    \vspace{-10pt}
\end{figure}

\vspace{-20pt}
\paragraph{Domain Adaption on Different Densities.}
To demonstrate the effectiveness of our DTS on overcoming the domain gap induced by LiDAR densities, we implement experimental comparison by adapting the nuScenes-trained model to different down-sampled KITTI dataset of different densities.
As nuScenes data is collected with 32-beam LiDAR, we down-sample KITTI data to 16-, 32-, 48-beams for simulation of different density situations, including the high-density to low-density or similar density, and the low-density to high-density.
As shown in Figure~\ref{fig:different_beam}, we compare our method with Source Only, SN \cite{wang2020train} and ST3D \cite{yang2019std} with the metric $\text{AP}_\text{{3D}}$.
We observe that, with the increase in density from 32-beam to 64-beam, the performance of ST3D has almost no improvement, while the performance of SN even becomes worse.
This is because SN and ST3D suffer from density gaps since their general detectors are sensitive to density of points without any special design.
Instead, our DTS trains density-insensitive detectors to overcome the density-induced domain gap, thus outperforming others in all density settings.

\vspace{-6pt}
\subsection{Ablation Studies}
\vspace{-6pt}
All ablation studies are conducted on nuScenes $\rightarrow$ KITTI, using SECOND-IoU as the network backbone.

\vspace{-8pt}
\paragraph{Main ablation.} 
As demonstrated in Table~\ref{table:ablations}, we investigate the contribution of each component.
Starting from the backbone model (a), we pre-train the 3D detector without RBRS in the source domain, then directly fine-tune the model through our basic teacher-student framework without RBRS and object graph alignment.  
By applying our proposed RBRS into the pre-training on the source data, the variant (b) brings significant improvement since it is beneficial to density-insensitivity.
Compared to (b), (c) does not introduce any boost as there is no annotation supervision in the target domain.
Moreover, by applying the RBRS and both NLC, ELC of the object graph alignment into the self-training, the performances of variants (d), (e), (f) improve a lot, demonstrating the effectiveness of each component.
\input{tables/ablation}

\vspace{-8pt}
\paragraph{Ablations on the re-sampling strategy.}
As in Table~\ref{tab:sampling}, we test different re-sampling strategies on both source only method and our DTS.
Since KITTI is denser than nuScenes, all strategies are designed to up-sample the nuScenes data or down-sample the KITTI data to obtain similar density distribution.
In the experiments for Source Only, we see that RBRS performs better than directly up-sampling the point cloud.
In the experiments for DTS, RBRS also achieves the best performance among different re-sampling strategies. 
We also observe that there is almost no improvement by simply up-sampling the source domain or down-sampling the target domain.
We think the reason is that a simple up-sampling or down-sampling strategy could neither introduce more information from data nor train a density-insensitive detector.

\input{tables/sample_methods}

\vspace{-8pt}
\paragraph{Ablations on the teacher-student framework.}
To investigate the effectiveness of our object-graph based teacher-student framework, we also compare our model with different self-training pipelines, like naive Teacher-Student \cite{xie2020self} and ST3D \cite{yang2021st3d}.
As shown in Table~\ref{tab:framework}, our object-graph based teacher-student framework outperforms the two compared self-training frameworks, since we explore the object consistency for learning density-invariant bounding boxes.

\input{tables/ST_framework}

\vspace{-8pt}
\paragraph{Sensitive analysis of node-level consistency and edge-level consistency.}
As shown in Table \ref{tab:sensitive}, we evaluate different $\beta_1$ and $\beta_2$ to control the weights of NLC and ELC. 
When $\beta_1$ and $\beta_2$ changed, the overall performance remains relatively good while $\text{AP}_\text{{3D}}$ is more sensitive to NLC. 
\input{tables/sensitive}


%% file: tables/datasets.tex
\begin{table}[t!]
\centering
\scalebox{0.80}{
\begin{tabular}{lccc}
\hline
Datasets & Size & LiDAR Type & Vertical Field  \\ \hline
Waymo    & 230K & $1 \times 64$ + $4 \times 200$-Beam    & $[-17.6^{\circ},2.4^{\circ}]$              \\
KITTI    & 15K & $1 \times 64$-Beam    & $[-23.6^{\circ},3.2^{\circ}]$             \\
nuScenes & 40K & $1 \times 32$-Beam    & $[-30.0^{\circ},10,0^{\circ}]$      \\ \hline      
\end{tabular}}
\vspace{-10pt}
\caption{Datasets overview. The dataset size refers to the number of annotated point cloud frames. }
\vspace{-20pt}
\label{tab:datasets}
\end{table}

%% file: tables/main_experiment.tex
\begin{table*}[ht]
\small
\centering
\setlength{\tabcolsep}{0.3mm}{
\begin{tabular}{c|l|cc|cc|cc}
\hline
\multicolumn{1}{l|}{\multirow{2}{*}{Task}} & \multirow{2}{*}{Method}              & \multicolumn{2}{c|}{SECOND-IoU\cite{yang2021st3d}}                  & \multicolumn{2}{c|}{PV-RCNN\cite{shi2020pv}}                    & \multicolumn{2}{c}{PointPillars\cite{lang2019pointpillars}}                \\ 
\multicolumn{1}{l|}{}                      &                                      & $\text{AP}_{\text{BEV}}/\text{AP}_{\text{3D}}$    & Closed Gap                 & $\text{AP}_{\text{BEV}}/\text{AP}_{\text{3D}}$   & Closed Gap                 & $\text{AP}_{\text{BEV}}/\text{AP}_{\text{3D}}$   & Closed Gap                 \\ \hline
\multirow{7}{*}{N$\rightarrow$K}         & Source Only                          & 51.8/17.9         & -/-                        & 68.2/37.2        & -/-                        & 22.8/0.5           & -/-                        \\
                                           & $\text{SN}^{\dag}$\cite{wang2020train}               & 59.7/37.6         & +25.1\%/+35.4\%  & 60.5/49.5        & +36.8\%/+27.1\%          & 39.3/2.0           & +26.6\%/+2.1\%           \\ \cline{2-8} 
                                           & ST3D\cite{yang2021st3d}               & 75.9/54.1         & +76.6\%/+59.5\%          & 78.4/70.9        & +49.0\%/+74.3\%          & 60.4/11.1          & +60.6\%/+14.9\%          \\
                                           & ST3D++\cite{yang2021st3d++}          & 80.5/ 62.4          & +91.1\%/+80.0\%          & -/-                & -/-                        & -/-                & -/-                        \\
                                           & 3D-CoCo\cite{yihan2021learning}      & -/-                 & -/-                        & -/-                & -/-                        & 77.0/47.2          & +87.4\%/+65.7\%          \\
                                           & Ours                                 & \textbf{81.4/66.6}  & \textbf{+94.0\%/+87.6\%} & \textbf{83.9/71.8} & \textbf{+75.8\%/+76.4\%} & \textbf{79.5/51.8} & \textbf{+91.5\%/+72.2\%} \\ \cline{2-8} 
                                           & Oracle                               & 83.3/73.5         & -/-                        & 88.9/82.5        & -/-                        & 84.8/71.6          & -/-                        \\ \hline
\multirow{7}{*}{W$\rightarrow$K}         & Source Only                          & 67.6/27.5         & -/-                        & 61.2/22.0        & -/-                        & 47.8/11.5          & -/-                        \\
                                           & $\text{SN}^{\dag}$\cite{wang2020train}               & 79.0/59.2         & +72.3\%/+69.0\%          & 79.8/63.6        & +66.9\%/+68.7\%          & 27.4/6.4           & -55.1\%/-8.5\%           \\ \cline{2-8} 
                                           & ST3D\cite{yang2021st3d}               & 82.2/61.8         & +93.0\%/+74.7\%          & 84.1/64.8        & +82.4\%/+70.7\%          & 58.1/23.2          & +27.8\%/+19.5\%          \\
                                           & ST3D++\cite{yang2021st3d++}          & 80.8/65.6         & +84.1\%/+82.8\%          & -/-                & -/-                        & -/-                & -/-                        \\
                                           & 3D-CoCo\cite{yihan2021learning}      & -/-                 & -/-                        & -/-                & -/-                        & \textbf{76.1}/42.9          & \textbf{+76.5\%}/+52.2\%          \\
                                           & Ours                                 & \textbf{85.8/71.5}  & \textbf{+115.9\%/+95.7\%} & \textbf{86.4/68.1} & \textbf{+90.6\%/+76.2\%}    & \textbf{76.1/50.2}          & \textbf{+76.5\%/+64.4\%}                          \\ \cline{2-8} 
                                           & Oracle                               & 83.3/73.5         & -/-                        & 89.0/82.5        & -/-                        & 84.8/71.6          & -/-                        \\ \hline
\multirow{8}{*}{W$\rightarrow$N}         & Source Only                          & 32.9/17.2         & -/-                        & 34.5/21.5        & -/-                        & 27.8/12.1        & -/-                        \\
                                           & $\text{SN}^{\dag}$\cite{wang2020train}               & 33.2/18.6         & +1.7\%/+7.5\%            & 34.2/22.3        & -1.5\%/+4.8\%            & 28.3/13.0        & +2.4\%/+4.7\%            \\ \cline{2-8} 
                                           & ST3D\cite{yang2021st3d}               & 35.9/20.2         & +15.9\%/+16.7\%          & 36.4/23.0        & +10.3\%/+8.8\%           & 30.6/15.6        & +13.2\%/+18.2\%          \\
                                           & ST3D++\cite{yang2021st3d++}          & 35.7/20.9         & +14.7\%/+20.9\%          & -/-                & -/-                        & -/-                & -/-                        \\
                                           & 3D-CoCo\cite{yihan2021learning}      & -/-                 & -/-                        & -/-                & -/-                        & 33.1/20.7          & +25.0\%/+44.8\%          \\
                                           & L$\cdot$ D \cite{wei2022lidar}       & 40.7/22.9         & +41.1\%/+32.2\%           & 43.3/25.6        & +47.3\%/+24.0\%          & 40.2/19.1         & +58.4\%/+36.5\%          \\
                                           & Ours                                 & \textbf{41.2/23.0}& \textbf{+43.7\%/+32.8\%}  & \textbf{44.0/26.2} & \textbf{+51.1\%/+27.5\%}                       & \textbf{42.2/21.5}                   & \textbf{+67.9\%/+49.0\%}                           \\ \cline{2-8} 
                                           & Oracle                               & 51.9/34.9         & -/-                        & 53.1/38.6        & -/-                        & 49.0/31.3          & -/-                        \\ \hline
\end{tabular}}
\vspace{-9pt}
\caption{Performance comparison of different methods on different domain adaptation tasks. $\dag$: SN is weakly supervised with target domain statistics. Source Only indicates that the model trained on the source dataset is directly tested on the target dataset. Oracle indicates that the model is trained with labeled target data. We report $\text{AP}_{\text{BEV}}$ and $\text{AP}_{\text{3D}}$ over 40 recall positions of the car category at IoU = 0.7.}
\vspace{-20pt}
\label{table:main}
\end{table*}

%% file: tables/ablation.tex
\begin{table}[t!]
\small
\centering
\scalebox{0.90}{
\begin{tabular}{c|c|ccc|cc}
\hline
  & Pre-Training & \multicolumn{3}{c|}{Self-Training} & \multirow{2}{*}{$\text{AP}_\text{{BEV}}$} & \multirow{2}{*}{$\text{AP}_\text{{3D}}$} \\ \cline{2-5}
      & RBRS         & RBRS             & NLC             &    ELC &                     &                        \\ \hline
(a)   &              &                  &                & & 73.3                    & 55.8                   \\
(b)   & \checkmark   &                  &                & & 77.6                    & 60.7                   \\
(c)   &              & \checkmark       &                & & 74.7                    & 60.7                   \\
(d)   & \checkmark   & \checkmark       &                & & 79.0                      & 61.7                   \\
(e)   & \checkmark   & \checkmark       & \checkmark     & & 79.5                      & 64.7                   \\
(f)   & \checkmark   & \checkmark       & \checkmark     & \checkmark & \textbf{81.4}                    & \textbf{66.6}                   \\ \hline
\end{tabular}
}
\vspace{-5pt}
\caption{Main ablations on the architecture design. NLC and ELC represent node-level consistency and edge-level consistency. }
\vspace{-10pt}
\label{table:ablations}
\end{table}

%% file: tables/sample_methods.tex
\begin{table}[t!]
\small
\centering
\scalebox{0.80}{
\begin{tabular}{l|ccccc|cc}
\hline
                                & \multicolumn{2}{c|}{Source Data}                              & \multicolumn{3}{c|}{Target Data}                                                       & \multicolumn{1}{c}{}                         & \multicolumn{1}{c}{}                        \\
\multirow{-2}{*}{Method}      & \multicolumn{1}{c}{Beam} & \multicolumn{1}{c|}{RBRS}       & \multicolumn{1}{c}{Beam} & \multicolumn{1}{c}{Point} & \multicolumn{1}{c|}{RBRS} & \multicolumn{1}{c}{\multirow{-2}{*}{$\text{AP}_\text{{BEV}}$}} & \multicolumn{1}{c}{\multirow{-2}{*}{$\text{AP}_\text{{3D}}$}} \\ \hline
                                &                             & \multicolumn{1}{c|}{}           &                             &                              &                           & 60.5                                         & 42.4                                        \\
                                & \checkmark                  & \multicolumn{1}{c|}{}           &                             &                              &                           & 70.9                                         & 48.2                                        \\
\multirow{-3}{*}{\makecell{Source \\ Only}} &                             & \multicolumn{1}{c|}{\checkmark} &                             &                              &                           &  \textbf{74.7}                               & \textbf{54.6}                               \\ \hline
                                &                             & \multicolumn{1}{c|}{}           &                             &                              &                           & 73.9                                         & 56.8                                        \\
                                & \checkmark                  & \multicolumn{1}{c|}{}           &                             &                              &                           & 74.2                                         & 56.4                                        \\
                                &                             & \multicolumn{1}{c|}{\checkmark} &                             &                              &                           & 80.9                                         & 64.0                                        \\
                                &                             & \multicolumn{1}{c|}{}           & \checkmark                  &                              &                           & 74.8                                         & 56.0                                        \\
                                &                             & \multicolumn{1}{c|}{}           &                             & \checkmark                   &                           & 74.1                                         & 54.4                                        \\
                                &                             & \multicolumn{1}{c|}{}           &                             &                              & \checkmark                & 80.2                                         & 61.9                                        \\ 
\multirow{-7}{*}{DTS}         &                             & \multicolumn{1}{c|}{\checkmark} &                             &                              & \checkmark                & \textbf{81.4}                                & \textbf{66.6}                               \\ \hline
\end{tabular}}
\vspace{-5pt}
\caption{Ablation on the re-sampling strategy. Here we take the case of transferring from low-density into high-density as example. 
The sampling strategies include beam-level up-sampling and RBRS on the source domain, as well as beam-level down-sampling, point-level down-sampling and RBRS on the target domain.}
\vspace{-5pt}
\label{tab:sampling}
\end{table}

%% file: tables/ST_framework.tex
\begin{table}[t!]
\small
\centering
\begin{tabular}{l|cc}
\hline
Self-Training Framework & \multicolumn{1}{l}{$\text{AP}_\text{{BEV}}$} & \multicolumn{1}{l}{$\text{AP}_\text{{3D}}$} \\ \hline
naive Teacher-Student\cite{xie2020self}                & 78.8                       & 55.5                          \\
ST3D\cite{yang2021st3d}                    & 78.7                       & 59.1                      \\
Ours                    & \textbf{81.4}                       & \textbf{66.6}                      \\ \hline
\end{tabular}
\vspace{-5pt}
\caption{Ablation on different self-training frameworks. All the frameworks are implemented with RBRS in both the pre-training stage and self-training stage.}
\vspace{-20pt}
\label{tab:framework}
\end{table}

%% file: tables/sensitive.tex
\begin{table}[h]
\centering
\scalebox{0.85}{
\begin{tabular}{
>{\columncolor[HTML]{EFEFEF}}c c|cc|c
>{\columncolor[HTML]{EFEFEF}}c |cc}
\hline
\multicolumn{1}{l}{\cellcolor[HTML]{EFEFEF}$\beta_1$} & \multicolumn{1}{l|}{$\beta_2$} & \multicolumn{1}{l}{$\text{AP}_\text{{BEV}}$} & \multicolumn{1}{l|}{$\text{AP}_\text{{3D}}$} & \multicolumn{1}{l}{$\beta_1$} & \multicolumn{1}{l|}{\cellcolor[HTML]{EFEFEF}$\beta_2$} & \multicolumn{1}{l}{$\text{AP}_\text{{BEV}}$} & \multicolumn{1}{l}{$\text{AP}_\text{{3D}}$} \\ \hline
0.05                                               & 0.0                         & 79.5                       & 64.7                         & 0.00                       & 0.3                                                 & 79.5                       & 62.7                      \\
0.05                                               & 0.1                         & 81.0                       & 63.9                         & 0.02                       & 0.3                                                 & 81.1                       & 64.0                      \\
0.05                                               & 0.2                         & 81.0                       & 64.3                         & 0.04                       & 0.3                                                 & 81.3                       & 64.5                      \\
0.05                                               & 0.3                         & \textbf{81.4}                       & \textbf{66.6}                         & 0.05                       & 0.3                                                 & \textbf{81.4}                       & \textbf{66.6}                      \\
0.05                                               & 0.4                         & 81.1                       & 64.1                         & 0.06                       & 0.3                                                 & 81.0                       & 64.0                      \\
0.05                                               & 0.5                         & 80.6                       & 61.5                        & 0.08                       & 0.3                                                 & 81.1                       & 63.8                      \\ \hline
\end{tabular}}
\vspace{-5pt}
\caption{Sensitivity analysis of NLC and ELC.}
\vspace{-15pt}
\label{tab:sensitive}
\end{table}